\documentclass[10pt,twocolumn,letterpaper]{article}

\usepackage[pagenumbers]{cvpr}

\usepackage{cuted}
\usepackage{xspace}
\usepackage{multirow}

\newcommand{\method}{\methodname\xspace}
\AtBeginDocument{\ifdefined\pdfstringdefDisableCommands\pdfstringdefDisableCommands{\let\method\methodname}\fi}

\newcommand{\Controller}{\textsc{Controller}\xspace}
\newcommand{\Locator}{\textsc{Locator}\xspace}
\newcommand{\Selector}{\textsc{Selector}\xspace}
\newcommand{\Verifier}{\textsc{Verifier}\xspace}
\newcommand{\aRefine}{\texttt{Refine}\xspace}
\newcommand{\aSwitch}{\texttt{Switch}\xspace}
\newcommand{\aUp}{\texttt{Up}\xspace}
\newcommand{\aInspect}{\texttt{Inspect}\xspace}

\DeclareMathOperator*{\argmax}{arg\,max}

\definecolor{cvprblue}{rgb}{0.21,0.49,0.74}
\usepackage[pagebackref,breaklinks,colorlinks,allcolors=cvprblue]{hyperref}

\def\confName{CVPR}
\def\confYear{2027}

\title{
 \method{}: Decoupling System One Search from System Two Answering \\ for Long Video Understanding
}

\author{%
  \bfseries Chengzhong Wang$^{1}$\thanks{Equal Contribution.} \quad
  Lang Lin$^{2}$\footnotemark[1] \quad
  Xinyang Liu$^{2}$ \quad
  Yueqin Yin$^{2}$ \\
  \bfseries Mingyuan Zhou$^{2}$ \quad
  Shengwei An$^{1}$\thanks{Corresponding author: \texttt{swan@vt.edu}} \\
  $^{1}$Virginia Tech \quad
  $^{2}$University of Texas at Austin \quad}

\begin{document}

\twocolumn[{%
  \renewcommand\twocolumn[1][]{#1}%
  \maketitle
  \begin{center}
    \centering
    \includegraphics[width=\linewidth]{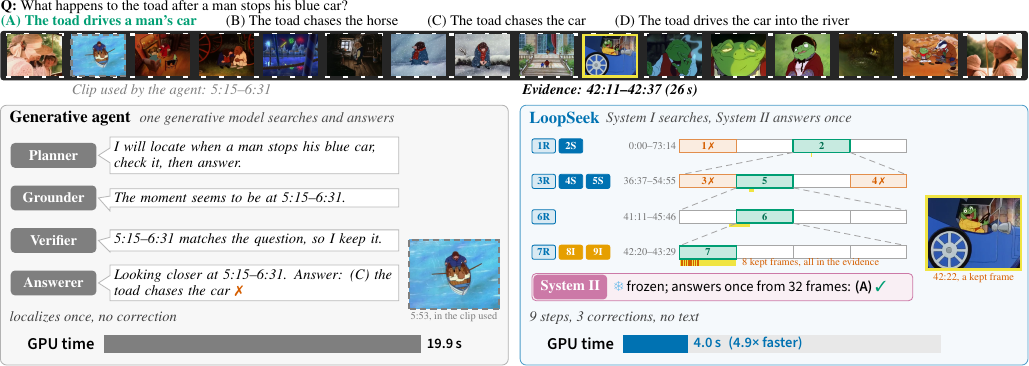}
    \captionof{figure}{\textbf{Decoupling search from answering.} On an LVBench question, a generative agent (left) reasons in text at every turn, localizes the moment once, at a clip about 36 minutes away from the evidence, and answers wrongly. \method (right) makes decisions without text: its search enters three wrong windows, switches away from each, reaches the evidence, and keeps 8 frames inside it; the frozen System~II then answers once. GPU time includes frame selection policy and answering.}
    \label{fig:teaser}
  \end{center}
}]
{\renewcommand{\thefootnote}{\fnsymbol{footnote}}%
\footnotetext[1]{Equal Contribution.}%
\footnotetext[2]{Corresponding author: \texttt{swan@vt.edu}}}

\begin{abstract}
Long video question answering requires finding relevant evidence before reasoning over it. With a limited frame budget, single-pass frame selection can miss critical evidence, while adaptive search often relies on repeated text generation to decide where to look. We introduce \method{}, a framework that decouples evidence search from answer generation through a division of labor between a non-thinking System I and a generative System II. Our key insight is that search can be expressed as a sequence of small, typed decisions, allowing System I to find the evidence that System II needs. Four System I role agents navigate a tree of video windows and coordinate through probability outputs to select frames, assess evidence, and revise the search without generating text. To align search with System II, System I learns from annotated evidence on questions that System II answers correctly only from that evidence, and the \Verifier's evidence estimates steer the search without calling System II, which stays frozen and answers once from the selected frames. On five long-video benchmarks, with a frozen Qwen3-VL-8B reading 32 frames, \method{} improves average accuracy from 53.5 to 63.9 over the uniform baseline and takes 32--80\% less GPU time per question than agentic methods. Project page: \url{https://aronwang02.github.io/LoopSeek/}
\end{abstract}

\suppressfloats[t]
\section{Introduction}
\label{sec:intro}

Language models think by generating text, and they pay for it at every step: even a small judgment requires sequential token generation.
System~I models~\cite{typesafe2026jev,onejev,chen2026system1arena} decouple judging from thinking.
Given a state and typed questions, such a model returns a calibrated probability for each option in one forward pass, without generating any text.
This enables a division of labor between a fast System~I that decides and a generative System~II that thinks, so that expensive reasoning is used only when necessary.
Recent work exploits this division by letting System~I take over decisions from System~II, either by replacing it at fixed decision points~\cite{barbosa2026pentest} or by answering first and escalating uncertain cases to it~\cite{demir2026commit,rahimi2026wireless,roitman2026guide}.
In both cases, System~I substitutes for System~II on decisions that System~II could also make.
While this reduces inference cost, it does not address a more fundamental limitation: what information System~II receives in the first place.\looseness=-1

This limitation is particularly severe in long-video understanding.
Under a limited frame budget, a System~II multimodal language model (MLLM) can examine only a sparse view of a long video.
If the selected frames miss the relevant evidence, no amount of reasoning can reliably recover the missing information.
Moreover, the evidence needed to answer a question is often concentrated within a short temporal interval~\cite{zhang2026vts}.
Uniform sampling at an affordable rate can easily miss such evidence, while adding frames raises computational cost and may bury relevant information among irrelevant content~\cite{yu2025framevoyager}.
Long-video understanding is therefore a search problem before it is an answering problem: the evidence must first be located under a fixed frame budget.

Existing approaches address this search problem in two main ways.
Frame selection methods~\cite{tang2025aks,liu2025bolt,yu2025framevoyager}
identify question-relevant frames and pass them to an answering model (System~II).
While efficient, these methods typically select frames without iterative feedback, making it difficult to recover from incorrect selections or verify whether the retrieved frames contain sufficient evidence.
Agentic methods~\cite{wang2024videoagent,zhang2025dvd,zhang2026vts}, in contrast, search over multiple turns and can revise earlier decisions.
However, they rely on a generative model for both searching and answering.
At each step, it examines frames or captions and reasons about where to look next, before performing further reasoning to produce the final answer.
We refer to this reliance on generative reasoning for both tasks as \emph{search--answer coupling}.
As a result, every search step incurs the cost of text generation, making inference increasingly expensive as the search progresses.
This raises a fundamental question: \emph{Can we decouple search from thinking, allowing a non-thinking System~I to find what System~II needs without calling it?}

The answer is positive because searching and answering require different kinds of computation.
Searching involves a sequence of relatively simple decisions, such as
whether this window/frame contains the evidence and
where to search next.
These decisions can be expressed as typed questions with a small number of options. Because they must be made repeatedly, their computational cost is critical.
Answering, by contrast, requires integrating the retrieved evidence and may involve more complex reasoning, such as counting objects or ordering events.
It is performed only once and therefore justifies more expensive computation.
The former is well suited to System~I, which can now also process visual and video inputs~\cite{onejev,jevomni,valen,zhu2026videojev}, while the latter is better handled by a generative System~II.

We therefore propose \method{}, which decouples search from answering for video understanding (\cref{fig:teaser}).
A non-thinking System~I carries out the entire search, while a frozen generative System~II answers from the frames System~I retrieves.
Unlike existing System~I--II collaborations, where System~I replaces System~II on selected decisions, \method{} assigns the two systems complementary roles in the same task.
The search is efficient by construction, since it generates no text and keeps System~II out of the loop, and System~II needs no retraining.
\method{} thus
combines the adaptivity of agentic search with the separation of search and answering offered by frame selection methods.

Realizing this separation introduces two key challenges.
First, how can System~I identify the evidence that System~II needs without querying it during search?
We address this through evidence-aligned training.
Specifically, System~I is trained on questions paired with annotated evidence, with System~II participating in the \emph{offline} selection of these questions.
Second, how can multiple System~I agents coordinate without free-form text?
We introduce four role agents that collaboratively navigate a hierarchical tree of video windows through structured probabilistic decisions.
The \Locator and \Selector identify candidate windows and frames, the \Verifier evaluates the retrieved evidence through a few judgments, and the \Controller uses these judgments to guide subsequent search actions.

Our contributions are summarized as follows:
\begin{itemize}[noitemsep,topsep=2pt,leftmargin=20pt]
  \item
    We introduce \method{}, a System~I--II framework that eliminates search--answer coupling by assigning search to a non-generative System~I and answering to a frozen System~II.
    Evidence-aligned training enables System~I to search independently for information relevant to System~II, without invoking it during inference.
  \item
    We develop a hierarchical search mechanism coordinated by four specialized System~I agents.
    Through structured probabilistic communication, these agents iteratively locate, evaluate, and refine evidence selection, enabling adaptive search without free-form text.
  \item
    Extensive experiments on 5 long-video benchmarks against 14 baselines demonstrate the effectiveness and efficiency of \method{}. Using Qwen3-VL-8B with a 32-frame budget, \method{} improves average accuracy from 53.5\% with uniform sampling to 63.9\%. It also achieves up to 4.9$\times$ speedup over the baselines.
\end{itemize}

\section{Related Work}
\label{sec:related}

\noindent
\textbf{Fast and Slow Systems.}
Dual-process theories describe cognition as the interplay of a fast, automatic System~I and a slow, deliberate System~II~\cite{kahneman2011thinking,evans2013dual}.
This division has been carried into AI to pair a cheap, reactive model with an expensive reasoner, in interactive agents~\cite{lin2023swiftsage}, language models~\cite{su2025dualformer,saha2025system1x}, and robot foundation models~\cite{shi2025hirobot,bjorck2025gr00t,bu2024robodual}.
Jev~\cite{typesafe2026jev} brings this division to foundation models: a System~I model answers typed questions about its input with calibrated probabilities over the options instead of generating text, leaving open-ended reasoning to System~II models such as large language models.
Open multimodal System~I models~\cite{onejev,jevomni,valen,zhu2026videojev} extend this idea to images and video.
We build our System~I on such open models and study how it should interact with a frozen, thinking System~II for long videos: System~I is a team of role agents that searches the video for the evidence that System~II needs, while System~II reasons once over what System~I curates.

\smallskip
\noindent
\textbf{Long Video Understanding with MLLMs.}
Video MLLMs~\cite{bai2025qwen25vl,bai2025qwen3vl,zhu2025internvl3,zhang2025llavavideo,wang2025internvideonext} typically answer from a fixed number of uniformly sampled frames.
Long-video benchmarks~\cite{fu2025videomme,wu2024longvideobench,zhou2025mlvu,wang2025lvbench,mangalam2023egoschema,chen2025cgbench,ye2025tstar} evaluate them on videos of minutes to hours.
To cover longer videos, recent work extends the visual context window~\cite{zhang2025longva} or compresses visual tokens across space and time~\cite{shen2025longvu}.
Rather than enlarging the input of the answering model, we keep its frame budget fixed and decide which frames fill it.

\smallskip
\noindent
\textbf{Keyframe Selection and Temporal Search.}
Query-aware frame selection scores candidate frames against the question with a vision-language model and keeps the most relevant ones while preserving temporal coverage~\cite{tang2025aks,liu2025bolt}, or learns which frame combinations help the answerer~\cite{yu2025framevoyager,lin2025glus}.
SeViLA~\cite{yu2023sevila} chains a keyframe localizer with an answerer derived from the same image-language model.
Temporal search methods cast long-video understanding as finding a few needle frames in a haystack~\cite{ye2025tstar}.
Most of these methods select frames in a single pass without checking that they show the evidence.
\method{} instead searches a window hierarchy over many turns: it moves between windows, inspects candidate frames, verifies whether each frame shows evidence, and judges where the remaining evidence lies.

\smallskip
\noindent
\textbf{Agentic Video Understanding.}
Another line relies on a large language model to reason over captions or summaries of video clips~\cite{zhang2024llovi,wang2025videotree}, or runs an agent loop in which the model plans, retrieves frames or captions with tools, and decides when it has gathered enough information~\cite{wang2024videoagent,fan2024videoagent,yang2025vca,zhang2025dvd,zhang2026vts}.
These systems place the large model in the loop as the planner, so every search step costs a call to it.
VideoMind~\cite{liu2026videomind} splits one model into planner, grounder, verifier, and answerer roles by switching LoRA adapters, which is the closest to our multi-agent design.
\method{} differs in where the roles sit and how they are learned.
Its planning, grounding, and verification agents form a lightweight System~I, learn by imitating an evidence expert, and communicate through compact conclusions.
The frozen System~II is called only once, to answer.

\section{Preliminaries}
\label{sec:prelim}

\noindent
\textbf{Problem Setup.}
We consider multiple-choice question answering over a long video $V$ of duration $T$.
A question $q$ comes with a set of options $\mathcal{O}$, one of which, $o^\star$, is correct.
System~II is a frozen MLLM answerer $A$ that reads a set $F$ of $K$ frames together with $q$ and generates its answer $A(F,q)$.
We write $p_A(o \mid F,q)$ for the probability that it assigns to option $o \in \mathcal{O}$.
System~I chooses $F$, with $|F|=K$, using at most $B$ actions, to help System~II answer correctly, \ie, $A(F,q)=o^\star$.
The common baseline instead samples the $K$ frames uniformly over $[0,T]$.
In both cases, System~II is called only once, on the final frame set.
System~I cannot check its choice during the search, since $o^\star$ is unknown and System~II is not called before the end.
It must therefore judge from the frames alone whether they show what System~II needs.
For training, each question is annotated with the key spans that contain its evidence, \eg, the clue intervals of CG-Bench~\cite{chen2025cgbench}.
We pad every span by a margin $\delta$ on both sides and merge overlapping spans into the evidence set $E \subset [0,T]$.
The evidence set is used only during training and is never observed at test time.

\smallskip
\noindent
\textbf{System~I Models.}
A System~I model~\cite{typesafe2026jev,onejev} answers typed questions about its input instead of generating free-form text.
Given an input $x$, made of text, frames, and a question with a finite option set $\mathcal{Y}=\{y_1,\dots,y_m\}$, it returns a distribution over $\mathcal{Y}$ in a single forward pass.
We label the options with letters and read the logits $z(x)$ of the letter tokens $c_1,\dots,c_m$ at the answer position,
\begin{equation}
  p_\theta(y_i \mid x) = \frac{\exp z_{c_i}(x)}{\sum_{j=1}^{m} \exp z_{c_j}(x)}.
  \label{eq:readout}
\end{equation}
Several questions about the same state can share one forward pass, each normalized over its own options.
A decision takes the most likely option or thresholds its probability, and the probabilities themselves can be passed on to other agents.
Every role agent of System~I takes this form, and we fine-tune each of them on the labeled option $y^\star$ with the cross-entropy over the options,
\begin{equation}
  \mathcal{L}(\theta) = -\log p_\theta(y^\star \mid x).
  \label{eq:ce}
\end{equation}

\smallskip
\noindent
\textbf{Imitation Learning with DAgger~\cite{ross2011dagger}.}
A sequential agent observes a state $s_t$ at step $t$ and picks an action $a_t$ from a finite set with its policy $\pi_\theta$.
We train such agents to imitate an expert $\pi^\star$ that reads the evidence set $E$.
In place of $E$, the student sees the frames and what the other agents predict.
Behavior cloning fits $\pi_\theta$ to the expert actions on the states that the expert visits.
At test time, however, the student's own mistakes lead it to states the expert never visits, where its errors compound~\cite{ross2011dagger}.
DAgger instead trains on the states that the student reaches.
In round $i$, it rolls out the mixture $\pi_i=\beta\,\pi^\star+(1-\beta)\,\pi_{\theta_i}$, which takes the expert action with probability $\beta$ at each step.
It labels each visited state $s$ with the expert action $\pi^\star(s)$, adds them to the aggregated set $\mathcal{D}$, and retrains the student on $\mathcal{D}$ with \cref{eq:ce}.

\section{Our Design of \method}
\label{sec:method}

\begin{figure*}[t]
\centering
\includegraphics[width=\linewidth]{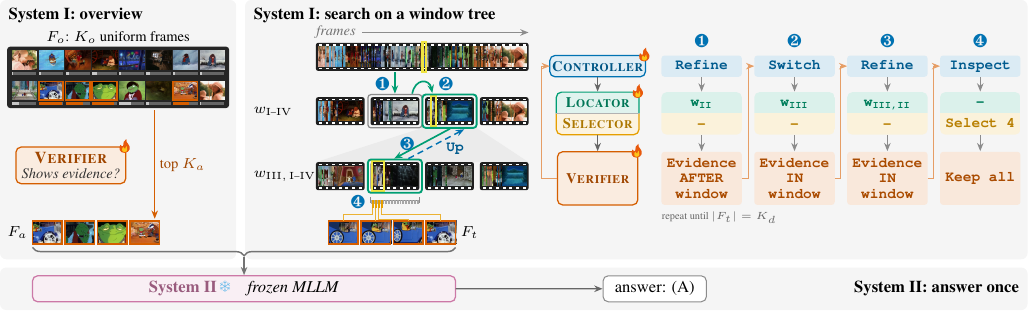}
\caption{\textbf{Overview of \method{}.} System~I judges the overview frames and searches a tree of windows over the video. At each step, the \Controller reads the \Verifier's judgments and the remaining budget and picks an action. The \Locator picks the next window, the \Selector picks frames in the active window, and the \Verifier keeps the frames that show evidence and judges where the evidence lies. After the search is complete, System~II answers once, from the selected overview frames and the kept frames.}
\label{fig:overview}
\end{figure*}

\method{} consists of System~I, which searches the video and gathers the evidence for the question, and System~II, a frozen answerer that answers once from the gathered frames (\cref{fig:overview}).
System~I has four role agents.
The \Controller plans the search, the \Locator and \Selector ground it, and the \Verifier checks what has been found.
They are the planning, grounding, and verification agents of \cref{sec:related}.
All four are System~I models (\cref{sec:prelim}), with parameters $\theta_c$ for the \Controller, $\theta_v$ for the \Verifier, and $\theta_g$ shared by both grounding roles.
Grounding yields fine-grained distributions, the \Verifier compresses them into a few judgments, and the \Controller decides from these judgments.
Every decision takes the most likely option under \cref{eq:readout}.

\subsection{Search on a Window Tree}
\label{sec:method_search}

\noindent
\textbf{Window Tree.}
A window is an interval $w=[w^-,w^+]\subseteq[0,T]$ of length $|w|=w^+-w^-$.
The root is $w_0=[0,T]$, and every window splits into $N$ equal children that do not overlap,
\begin{equation}
  \mathrm{ch}(w)=\big\{[\,w^-+\tfrac{k-1}{N}|w|,\,w^-+\tfrac{k}{N}|w|\,]\big\}_{k=1}^{N}.
  \label{eq:children}
\end{equation}
Every window $w\neq w_0$ thus has a unique parent $\mathrm{pa}(w)$, the window that it was split from, \ie, $w\in\mathrm{ch}(\mathrm{pa}(w))$.
Its depth is $\ell(w)=\ell(\mathrm{pa}(w))+1$, with $\ell(w_0)=0$.
We also write $\tilde w=[w^--\mu|w|,\,w^++\mu|w|]\cap[0,T]$ for the window $w$ extended by a margin $\mu|w|$ on both sides.

\smallskip
\noindent
\textbf{Overview.}
System~I first samples a set $F_{o}$ of $K_o$ overview frames uniformly over $[0,T]$.
The \Verifier judges every overview frame, and the $K_a$ frames that most likely show the evidence form the answer overview $F_{a}\subseteq F_{o}$.
$F_{a}$ goes to System~II, and the search gathers what it misses.

\smallskip
\noindent
\textbf{Search.}
The search state $s_t$ holds the active window $w_t$, the visited windows $\mathcal{W}_t$, the kept and rejected frames $F_t$ and $R_t$, and the number $b_t$ of remaining actions, starting from $w_0$, $F_0=R_0=\emptyset$, and $b_0=B$.
At step $t$, the \Controller picks a legal action $a_t\in\mathcal{A}(s_t)\subseteq\{\aRefine,\aSwitch,\aUp,\aInspect\}$, and
{\small
\begin{equation}
  w_{t+1}=
  \begin{cases}
    \Locator\big(\mathrm{ch}(w_t)\setminus\mathcal{W}_t\big) & \text{if } a_t=\aRefine,\\
    \Locator\big(\mathrm{ch}(\mathrm{pa}(w_t))\setminus\mathcal{W}_t\big) & \text{if } a_t=\aSwitch,\\
    \mathrm{pa}(w_t) & \text{if } a_t=\aUp,\\
    w_t & \text{if } a_t=\aInspect,
  \end{cases}
  \label{eq:transition}
\end{equation}
}
where $\Locator(\mathcal{C}_t)$ is the window that the \Locator picks (\cref{sec:method_roles}) among the candidates $\mathcal{C}_t$, the children of $w_t$ or of $\mathrm{pa}(w_t)$ that are not yet visited, and \aInspect also updates $F_t$ and $R_t$ by \cref{eq:keep}.
\aUp needs $\ell(w_t)\ge2$, since returning to the root and refining again would duplicate \aSwitch.
The search always begins with \aRefine and stops at the first step $t_{\mathrm{end}}$ with $|F_t|=K_d$, $b_t=0$, or $\mathcal{A}(s_t)=\emptyset$.
If fewer than $K_d$ frames are kept, the remaining slots are filled from $R_{t_{\mathrm{end}}}$ in decreasing order of the \Verifier's evidence probability, and then from a uniform grid over the video.
System~II answers once from the resulting $K=K_a (\text{overview})+K_d (\text{detail})$ frames.

\subsection{Role Agents in System~I}
\label{sec:method_roles}

\begin{table}[t]
  \centering
  \footnotesize
  \setlength{\tabcolsep}{3pt}
  \begin{tabular}{@{}lp{0.65\linewidth}l@{}}
    \toprule
    Role & Typed question (options) & Label \\
    \midrule
    \Controller & Next action? (legal actions) & $\pi^\star$ \\
    \Locator & Which window holds the evidence? ($\mathcal{C}_t$) & $E$ \\
    \Selector & Which frames show the evidence? ($c_1,\dots,c_M$) & $E$ \\
    \Verifier & Does frame $c$ show the evidence? (yes, no) & $E$ \\
              & Where is the evidence? (four positions) & $E$ \\
              & Nearest ancestor with unvisited evidence? (ancestors, none) & $E$ \\
    \bottomrule
  \end{tabular}
  \caption{\textbf{Role agents.} Roles answer typed questions by \cref{eq:readout}. All labels come from the evidence set $E$, directly or via expert $\pi^\star$.}
  \label{tab:roles}
\end{table}

Each agent answers typed questions with \cref{eq:readout} (\cref{tab:roles}).
We use $x_t$ to denote the input of an agent at step $t$, which always starts with the question and its options. The inputs of the \Controller and the \Verifier continue with $F_{o}$.\looseness=-1

\smallskip
\noindent
\textbf{\Locator.}
Given the candidate windows $\mathcal{C}_t$ in \cref{eq:transition}, \Locator selects the window with the highest score: $\Locator(\mathcal{C}_t)=\argmax_{u\in\mathcal{C}_t}p_{\theta_g}(u\mid x_t)$.
Each candidate is represented by its temporal span and context frames inside it.
The context frames are $F_{o}$ near the root, and deeper in the tree they are sampled uniformly from the parent of the candidates, so that every candidate gets the same number of frames.
For \aSwitch, $x_t$ also includes the \Verifier's judgment on whether the evidence lies before or after $w_t$.

\smallskip
\noindent
\textbf{\Selector.}
\aInspect draws $M$ candidate frames $c_1<\dots<c_M$ uniformly from $\tilde w_t$, excluding frames picked before.
\Selector scores all candidates in one forward pass, $\sigma_j=p_{\theta_g}(c_j\mid x_t)$, and greedily selects $m$ frames in descending order of score, skipping the neighbors of earlier selections,
\begin{equation}
  j_i=\argmax_{j\in\mathcal{J}_i}\sigma_j,\quad
  \mathcal{J}_i=\{j: |j-j_{i'}|>1\ \forall\, i'<i\},
  \label{eq:select}
\end{equation}
for $i=1,\dots,m$, giving $S_t=\{c_{j_1},\dots,c_{j_m}\}$.

\smallskip
\noindent
\textbf{\Verifier.}
The \Verifier uses the question, overview frames $F_{o}$, and retained frames $F_t$ to make three types of judgments:
\emph{1) Frame evidence.}
In one forward pass, with one yes/no question per frame,
it estimates the probability that each queried frame $c$ contains relevant evidence:
$v(c)=p_{\theta_v}(\text{yes}\mid x_t,c)$.
For the overview frames, it
selects the $K_a$ highest-scoring frames to form $F_a$.
For the selected frames $S_t$, it retains or rejects each frame with threshold $\kappa$:
\begin{equation}
  \begin{aligned}
    F_{t+1}&=F_t\cup\{c\in S_t: v(c)\ge\kappa\},\\
    R_{t+1}&=R_t\cup\{c\in S_t: v(c)<\kappa\},
  \end{aligned}
  \label{eq:keep}
\end{equation}
subject to a maximum of $K_d$ retained frames.
\emph{2) Evidence position.}
From frames sampled uniformly from $\tilde w_t$, it predicts whether the evidence lies before $\tilde w_t$, inside it while covering less than a fraction $\rho$ of it, inside it while covering more, or after it.
\emph{3) Ancestor.}
For $\ell(w_t)\ge2$, it predicts the smallest ancestor of $w_t$ that still holds evidence in windows not yet visited, or none.
The \Verifier judges the overview frames once, and the picked frames after every inspection.
It predicts the evidence position after every inspection and, outside the root, before every \Controller decision for which $w_t$, $F_t$, or the inspections in $w_t$ have changed.

\smallskip
\noindent
\textbf{\Controller.}
The \Controller picks $a_t=\argmax_{a\in\mathcal{A}(s_t)}p_{\theta_c}(a\mid x_t)$.
Its input describes $s_t$ in text: the active and visited windows, $\ell(w_t)$, the number of inspections in $w_t$, how many of them kept fewer than $m/2$ frames, how many frames the last one kept, $|F_t|$, and $b_t$.
It also holds $F_t$ and the latest \Verifier messages, which give the evidence position and the ancestor as probabilities.

\smallskip
\noindent
\textbf{Efficiency.}
Each System~I decision requires one forward pass without text generation.
For the inputs that hold $F_{o}$, the prefix formed by the task, the question, the options, and $F_{o}$ is fixed within an episode, so its key-value cache is computed once per role and reused later.
System~II is called once per question, after the search is complete.

\subsection{Learning from an Evidence Expert}
\label{sec:method_training}

\noindent
\textbf{Evidence that System~II needs.}
Since System~I never invokes System~II during search,
it must learn to locate what System~II needs from the evidence alone.
Following VTS~\cite{zhang2026vts}, we select training questions for which System~II answers correctly using frames within the annotated evidence $E$, but not from the question alone or frames outside $E$.
All training labels are then derived from $E$, aligning System~I's search with the evidence used by System~II.

\smallskip
\noindent
\textbf{Labels.}
Let $d(u)=|E\cap u|/|u|$ be the fraction of an interval $u$ covered by $E$.
For the \Locator, we label all candidates whose evidence coverage is at least half of the maximum:
\begin{equation}
  \mathcal{Y}^\star_{\Locator}=\big\{u\in\mathcal{C}_t: d(u)\ge\tfrac12\max_{u'\in\mathcal{C}_t}d(u')\big\},
  \label{eq:locator_label}
\end{equation}
and a decision with $\max_{u}d(u)=0$ receives no label.
The correct candidates of the \Selector are those in $E$, and a frame $c$ is labeled yes for the \Verifier if $c\in E$.
The evidence position and the ancestor are determined from $E$, $\tilde w_t$, and $\mathcal{W}_t$.
When multiple options are correct, we maximize their combined probability,
\begin{equation}
  \mathcal{L}(\theta)=-\log\sum_{y\in\mathcal{Y}^\star}p_\theta(y\mid x),
  \label{eq:setce}
\end{equation}
which reduces to \cref{eq:ce} for a single correct option.

\smallskip
\noindent
\textbf{Evidence Expert.}
We train \Controller to imitate an expert $\pi^\star$ with access to $E$.
Let $n_t$ denote the number of inspections in $w_t$, $n^-_t$ the number of them that kept fewer than $m/2$ frames, and $k_t$ the number of frames kept by the last one.
Then, the expert follows
\begin{equation}
  \pi^\star(s_t)=
  \begin{cases}
    \aRefine & \text{if } w_t=w_0,\\
    \mathrm{move} & \text{if } E\cap\tilde w_t=\emptyset,\\
    \mathrm{narrow} & \text{if } n_t=0 \text{ or } k_t\ge m/2,\\
    \aInspect & \text{if } n^-_t=1,\\
    \mathrm{deepen} & \text{otherwise}.
  \end{cases}
  \label{eq:expert}
\end{equation}
Here, $\mathrm{move}$ chooses \aSwitch if an unvisited sibling of $w_t$ meets $E$ and \aUp otherwise.
Similarly, $\mathrm{narrow}$ selects \aRefine if $d(\tilde w_t)<\rho$ and an unvisited child of $w_t$ meets $E$, and \aInspect otherwise.
Finally, $\mathrm{deepen}$ gives \aRefine if $\mathrm{narrow}$ does or if, with $b_t\ge2$, an unvisited child still has never-picked candidate frames in $E$, and $\mathrm{move}$ otherwise.
States that cannot reach $E$ are excluded from training.

\smallskip
\noindent
\textbf{Staged Training.}
To reduce the gap between training and inference, we train each role on progressively more realistic search states, initializing each from the base System~I model.
(i)~The expert controls the search, while the grounding roles either follow the annotations or act zero-shot, and the \Verifier receives annotation-derived messages.
The resulting states are used to train $\theta_g$ and $\theta_v$.
(ii)~We roll out $\pi^\star$ with the trained grounding roles and \Verifier, without any oracle, and retrain $\theta_g$ and $\theta_v$ on the new states mixed with those from the first stage.
(iii)~We roll out $\pi^\star$ with the final $\theta_g$ and $\theta_v$ and train $\theta_c$ by behavior cloning. \Controller receives predicted \Verifier messages rather than oracle annotations.
(iv)~We refine System~I with DAgger (\cref{sec:prelim}): the student searches, the expert and the annotations label the states that it reaches, and we retrain on the aggregated states.
The expert never enters a window without evidence, so in stages (ii) to (iv) we force some \Locator calls into such a window, which is then inspected once. The forced decisions get no labels, but the states after them train the \Verifier to recognize such windows and the \Controller to leave them (App. \ref{sec:supp_training}).\looseness=-1

\section{Experiments}
\label{sec:experiments}

\subsection{Setup}
\label{sec:exp_setup}

\noindent
\textbf{Benchmarks and Models.}
We report multiple-choice accuracy on Video-MME~\cite{fu2025videomme} without subtitles (2,700 questions), MLVU~\cite{zhou2025mlvu} dev (2,174 questions, averaged over its seven task types), LongVideoBench~\cite{wu2024longvideobench} val (1,337 questions), LVBench~\cite{wang2025lvbench} (1,549 questions), and CG-Bench~\cite{chen2025cgbench} mini (3,000 questions).
For System~II, we test five frozen MLLMs, Qwen3-VL-8B~\cite{bai2025qwen3vl}, Qwen2.5-VL-7B~\cite{bai2025qwen25vl}, Qwen2-VL-7B~\cite{Qwen2-VL}, LLaVA-OneVision-7B~\cite{li2024llava}, and InternVL3-8B~\cite{zhu2025internvl3}. They answer once from $K=32$ frames that System~I selects, which are independent of System~II.
System~I consists of three LoRA adapters~\cite{hu2022lora} on a frozen OneJev-4B~\cite{onejev}: $\theta_c$ for the \Controller, $\theta_v$ for the \Verifier, and $\theta_g$ shared by the \Locator and the \Selector.
At test time, $K_o=96$, $K_a=24$, $K_d=8$, and $B=24$.

\smallskip
\noindent
\textbf{Baselines.}
We compare our approach against 14 baselines: ReFoCUS~\cite{lee2025refocus}, VideoITG~\cite{wang2026videoitg}, ReMem~\cite{meng2026remem}, FOCUS~\cite{zhu2025focus}, BOLT~\cite{liu2025bolt}, SVAgent~\cite{yang2026svagent}, RACER~\cite{huang2026racer}, T*~\cite{ye2025tstar}, A.I.R.~\cite{zou2026air}, VideoMind~\cite{liu2026videomind}, Video-MTR~\cite{xie2026videomtr}, Conan-7B~\cite{ouyang2025conan}, VTS~\cite{zhang2026vts} and VideoMiner~\cite{cao2025videominer}.

\smallskip
\noindent
\textbf{Training.}
We train System~I on 5,025 questions from the training data of VideoITG~\cite{wang2026videoitg}, 1,530 outside the mini set of CG-Bench, and 1,493 from that of LV-Haystack~\cite{ye2025tstar}, all kept by the filter of \cref{sec:method_training} with Qwen3-VL-8B as System~II, and run one round of DAgger (\cref{sec:supp_training}).
During training, System~II reads $K_o=K_a=16$ uniform overview frames and $K_d=16$ searched frames.
We take the overview uniformly instead of picking it with the \Verifier as at test time, since a picked overview would hold much of the evidence and leave System~I too little to learn about locating detail frames.
The equal split also balances the roles of overview and searched frames, which lets the same System~I serve all splits in \cref{tab:split}.

\subsection{Main results}
\label{sec:exp_main}

\begin{table}[t]
  \centering
  \scriptsize
  \colorlet{singlepass}{black!55}
  \setlength{\tabcolsep}{2.5pt}
  \begin{tabular}{@{}llcccccc@{}}
    \toprule
    Method & Frames & V-MME & MLVU & LVBench & LongVB & CG & Avg. \\
    \midrule
    \multicolumn{8}{@{}l}{\textit{Qwen3-VL-8B~\cite{bai2025qwen3vl}}} \\
    \textcolor{singlepass}{Uniform$^{\dagger}$} & \textcolor{singlepass}{32} & \textcolor{singlepass}{63.9} & \textcolor{singlepass}{63.7} & \textcolor{singlepass}{39.1} & \textcolor{singlepass}{60.4} & \textcolor{singlepass}{40.7} & \textcolor{singlepass}{53.5} \\
    ReFoCUS$^{\dagger}$~\cite{lee2025refocus} & 32 & 68.5 & 72.5 & 49.3 & 63.3 & \underline{48.1} & 60.3 \\
    VideoITG$^{\dagger}$~\cite{wang2026videoitg} & 32 & \textbf{68.9} & 75.8 & \underline{53.5} & 63.6 & 47.3 & \underline{61.8} \\
    ReMem$^{\dagger}$~\cite{meng2026remem} & 32 & \underline{68.8} & \underline{77.6} & 53.3 & \underline{64.4} & -- & -- \\
    SVAgent~\cite{yang2026svagent} & $\le$32 & 63.1 & -- & 40.8 & 61.0 & -- & -- \\
    RACER~\cite{huang2026racer} & 32 & 67.9 & 73.5 & -- & 64.0 & -- & -- \\
    \method{} (ours) & 32 & \textbf{68.9} & \textbf{77.7} & \textbf{54.0} & \textbf{67.5} & \textbf{51.4} & \textbf{63.9} \\
    \midrule
    \multicolumn{8}{@{}l}{\textit{Qwen2.5-VL-7B~\cite{bai2025qwen25vl}}} \\
    \textcolor{singlepass}{Uniform$^{\dagger}$} & \textcolor{singlepass}{32} & \textcolor{singlepass}{62.4} & \textcolor{singlepass}{60.8} & \textcolor{singlepass}{37.8} & \textcolor{singlepass}{57.9} & \textcolor{singlepass}{35.4} & \textcolor{singlepass}{50.9} \\
    T*~\cite{ye2025tstar} & 32 & 56.3 & 58.1 & 37.7 & 57.4 & 35.8 & 49.1 \\
    A.I.R.~\cite{zou2026air} & $\le$32 & \textbf{65.0} & \underline{67.5} & 42.3 & \underline{61.4} & \underline{44.6} & \underline{56.2} \\
    VideoMind~\cite{liu2026videomind} & $\le$32 & 60.7 & 64.7 & \underline{45.2} & 57.6 & 39.4 & 53.5 \\
    Video-MTR$^{\diamond}$~\cite{xie2026videomtr} & $\le$32 & 59.0 & 58.0 & 38.2 & 52.3 & 33.7 & 48.3 \\
    Conan-7B$^{\diamond}$~\cite{ouyang2025conan} & 32 & 60.5 & 61.6 & 39.2 & 56.6 & 39.1 & 51.4 \\
    SVAgent~\cite{yang2026svagent} & $\le$32 & 61.2 & -- & 40.6 & 60.7 & -- & -- \\
    \method{} (ours) & 32 & \underline{64.7} & \textbf{69.2} & \textbf{49.7} & \textbf{65.3} & \textbf{47.3} & \textbf{59.3} \\
    \midrule
    \multicolumn{8}{@{}l}{\textit{Qwen2-VL-7B~\cite{Qwen2-VL}}} \\
    \textcolor{singlepass}{Uniform$^{\dagger}$} & \textcolor{singlepass}{32} & \textcolor{singlepass}{58.0} & \textcolor{singlepass}{60.3} & \textcolor{singlepass}{39.3} & \textcolor{singlepass}{55.3} & \textcolor{singlepass}{32.2} & \textcolor{singlepass}{49.0} \\
    FOCUS$^{\dagger}$~\cite{zhu2025focus} & 32 & 59.7 & 65.4 & \underline{47.5} & \underline{62.3} & 38.9 & \underline{54.8} \\
    A.I.R.~\cite{zou2026air} & $\le$32 & \underline{60.0} & \underline{66.6} & 45.6 & 58.9 & \underline{42.0} & 54.6 \\
    VideoMiner~\cite{cao2025videominer} & 32 & 58.4 & 65.1 & -- & 56.4 & -- & -- \\
    VideoMind~\cite{liu2026videomind} & $\le$32 & 58.0 & 62.7 & 40.8 & 56.3 & 38.4 & 51.2 \\
    \method{} (ours) & 32 & \textbf{63.6} & \textbf{70.5} & \textbf{49.1} & \textbf{62.7} & \textbf{45.7} & \textbf{58.3} \\
    \midrule
    \multicolumn{8}{@{}l}{\textit{LLaVA-OneVision-7B~\cite{li2024llava}}} \\
    \textcolor{singlepass}{Uniform$^{\dagger}$} & \textcolor{singlepass}{32} & \textcolor{singlepass}{58.4} & \textcolor{singlepass}{63.1} & \textcolor{singlepass}{39.3} & \textcolor{singlepass}{56.5} & \textcolor{singlepass}{31.4} & \textcolor{singlepass}{49.8} \\
    BOLT$^{\dagger}$~\cite{liu2025bolt} & 32 & 59.9 & 66.8 & 42.7 & 59.6 & 35.6 & 52.9 \\
    FOCUS$^{\dagger}$~\cite{zhu2025focus} & 32 & 58.3 & 64.9 & \underline{45.9} & 60.7 & 35.1 & 53.0 \\
    ReFoCUS$^{\dagger}$~\cite{lee2025refocus} & 32 & \textbf{62.6} & 68.5 & 45.8 & \underline{61.0} & \underline{37.6} & \underline{55.1} \\
    A.I.R.~\cite{zou2026air} & $\le$32 & 61.4 & \underline{69.3} & 45.1 & 60.7 & 37.1 & 54.7 \\
    \method{} (ours) & 32 & \underline{62.3} & \textbf{70.6} & \textbf{47.6} & \textbf{62.2} & \textbf{38.7} & \textbf{56.3} \\
    \midrule
    \multicolumn{8}{@{}l}{\textit{InternVL3-8B~\cite{zhu2025internvl3}}} \\
    \textcolor{singlepass}{Uniform$^{\dagger}$} & \textcolor{singlepass}{32} & \textcolor{singlepass}{64.9} & \textcolor{singlepass}{68.8} & \textcolor{singlepass}{43.9} & \textcolor{singlepass}{58.7} & \textcolor{singlepass}{42.7} & \textcolor{singlepass}{55.8} \\
    ReFoCUS$^{\dagger}$~\cite{lee2025refocus} & 32 & \underline{67.0} & 72.7 & 50.0 & 62.0 & \underline{48.1} & 60.0 \\
    A.I.R.~\cite{zou2026air} & $\le$32 & \textbf{68.2} & \textbf{74.5} & \underline{50.3} & \underline{62.8} & 47.9 & \underline{60.7} \\
    VideoMind~\cite{liu2026videomind} & $\le$32 & 64.2 & 66.5 & 42.3 & 59.7 & 43.0 & 55.1 \\
    \method{} (ours) & 32 & 65.4 & \underline{74.1} & \textbf{51.0} & \textbf{64.3} & \textbf{50.1} & \textbf{61.0} \\
    \bottomrule
  \end{tabular}
  \caption{\textbf{Accuracy (\%) grouped by answerer.}
    \textit{Italic}: the answerer, which answers from the frames that each method in its group selects.
    $^{\dagger}$: picks frames in a single pass. Unmarked methods search over multiple rounds, iteratively or as agents.
    Frames: number of frames that the answerer reads. Avg.: mean of the five benchmarks.
    --: not reported in the paper, and the code or weights are not released.
    V-MME: Video-MME; LongVB: LongVideoBench; CG: CG-Bench.
    ReMem parses the question with GPT-4o. $^{\diamond}$: answers with a model fine-tuned from the answerer.
    \textbf{Bold}: best in group, ties included; \underline{Underline}: second best. }
  \label{tab:main}
\end{table}

\noindent
\textbf{Accuracy.}
\cref{tab:main} reports the results grouped by answerer.
\method{} achieves the highest average accuracy across five benchmarks for every answerer, outperforming uniform sampling by up to 10.4 percentage points and the strongest baseline by up to 3.5 points.
It achieves the best or tied-best accuracy in 21 of the 25 answerer--benchmark combinations and outperforms all compared agentic methods on every benchmark for which their results are reported.
The improvements over uniform sampling are particularly pronounced on LVBench, CG-Bench, and MLVU, reaching up to 14.9 points, but are smaller on Video-MME (up to 5.6 points), which includes questions requiring holistic video understanding.
For Qwen3-VL-8B, \method{} improves accuracy by up to 8.8 points on questions about specific objects, actions, or text, while matching uniform sampling on video-level summarization questions.
Notably, although System~I is trained using Qwen3-VL-8B as System~II, its improvements transfer to four unseen answerers, suggesting that the learned search policy identifies broadly useful evidence rather than overfitting to a particular MLLM.

\begin{table}[t]
  \centering
  \footnotesize
  \begin{tabular}{@{}lcccc@{}}
    \toprule
    \multirow{2}{*}{Method} & \multicolumn{2}{c}{MLVU} & \multicolumn{2}{c}{LVBench} \\
    & Search & Total & Search & Total \\
    \midrule
    \method{} (ours) & \textbf{2.59} & \textbf{3.41} & \textbf{3.35} & \textbf{3.96} \\
    VideoMind-7B~\cite{liu2026videomind} & 3.63 & 5.04 & 4.54 & 6.13 \\
    VTS-8B~\cite{zhang2026vts} & 9.98 & 11.16 & 18.73$^{a}$ & 19.92$^{a}$ \\
    Conan-7B~\cite{ouyang2025conan} & -- & 16.51 & -- & 16.13 \\
    ReFoCUS-1.3B~\cite{lee2025refocus} & 4.42 & -- & 4.38 & -- \\
    \bottomrule
  \end{tabular}
  \caption{\textbf{GPU seconds per question.}
    One H200, batch size 1, the same 1000 questions per benchmark ($^{a}$218).
    Search: work before the answer, with vision encoding. Total: search and answer.}
  \label{tab:efficiency}
\end{table}

\smallskip
\noindent
\textbf{Efficiency}
\label{sec:exp_efficiency}
\cref{tab:efficiency} reports GPU time per question on the same randomly sampled questions from MLVU and LVBench, excluding CPU video decoding (App.~\ref{sec:supp_timing}).
Compared with VideoMind, VTS, and Conan, \method{} reduces GPU time by up to 35\%, 80\%, and 79\%, respectively.
Its search alone is also faster than the single-pass frame selector ReFoCUS.
System~I makes an average of 16.7 decisions per question, each requiring only one forward pass through a 4B-parameter model without text generation, while System~II is invoked only once for answering.

\smallskip
\noindent
\textbf{Evidence hit rate}
\label{sec:exp_evidence}
Accuracy alone does not directly show whether System~I found the evidence, so we also measure the evidence hit rate: the share of the frames given to System~II that lie in the annotated evidence, averaged over questions.
We use the clue intervals of CG-Bench mini and the time references of LVBench.
The evidence is short, with a median of 10\,s on CG-Bench and 17\,s on LVBench and a median under 1\% of the video.
Uniform frames land in it only in proportion to its length, so their hit rate stays at 1.4--1.5\% on CG-Bench and 9.6\% on LVBench from 32 to 128 frames (\cref{tab:evidence}) and adding uniform frames does not raise it.
With only 32 frames, \method{} achieves hit rates of 9.85\% and 15.20\%, representing 6.7$\times$ and 1.6$\times$ improvements over uniform sampling.
Most of the gain comes from the search: 26.34\% and 25.07\% of the 8 searched frames fall in the evidence, versus 4.35\% and 11.90\% of the 24 overview frames, which preserve broader video coverage.

The \Controller also enables recovery from unproductive search directions.
It takes an average of 6.4 actions per question, with 23.5\% of questions involving at least one \aSwitch or \aUp action.
For these questions, the proportion of windows containing evidence increases from 22.8\% before the first such action to 35.0\% afterward on CG-Bench, and from 20.1\% to 31.0\% on LVBench, an approximately 1.5$\times$ improvement on both benchmarks.

\begin{table}[t]
  \centering
  \footnotesize
  \begin{tabular}{@{}lccc@{}}
    \toprule
    Method & Frames & CG-Bench & LVBench \\
    \midrule
    Uniform & 32 & 1.46 & 9.61 \\
    Uniform & 64 & 1.44 & 9.60 \\
    Uniform & 128 & 1.45 & 9.59 \\
    \midrule
    \method{} (ours) & 32 & \textbf{9.85} & \textbf{15.20} \\
    \quad searched frames & 8 & 26.34 & 25.07 \\
    \quad overview frames & 24 & 4.35 & 11.90 \\
    \bottomrule
  \end{tabular}
  \caption{\textbf{Evidence hit rate (\%).} Share of a question's frames that lie inside the annotated evidence, averaged over questions. The last two rows split the 32 frames of \method{} into the 8 kept by the search and the 24 picked from the overview.}
  \label{tab:evidence}
\end{table}

\subsection{Ablations}
\label{sec:exp_ablation}

\begin{table}[t]
  \centering
  \footnotesize
  \setlength{\tabcolsep}{3pt}
  \begin{tabular}{@{}cc|cccccc@{}}
    \toprule
    $K_a$ & $K_d$ & V-MME & MLVU & LVBench & LongVB & CG & Avg. \\
    \midrule
    8 & 24 & 67.1 & 74.2 & 53.6 & 67.0 & \textbf{52.4} & 62.9 \\
    16 & 16 & 68.5 & 76.5 & 52.7 & 67.4 & 52.0 & 63.4 \\
    24 & 8 & \textbf{68.9} & \textbf{77.7} & \textbf{54.0} & \textbf{67.5} & 51.4 & \textbf{63.9} \\
    \bottomrule
  \end{tabular}
  \par\smallskip
  \begin{tabular}{@{}cc|cccccc@{}}
    \toprule
    \multirow{2}{*}{$K_a$} & \multirow{2}{*}{$K_d$} & \multicolumn{3}{c}{LVBench} & \multicolumn{3}{c}{CG-Bench} \\
    \cmidrule(lr){3-5}\cmidrule(l){6-8}
    & & $<1\%$ & 1--10\% & $\geq 10\%$ & $<1\%$ & 1--10\% & $\geq 10\%$ \\
    \midrule
    8 & 24 & 53.4 & \textbf{51.8} & 57.7 & \textbf{51.6} & \textbf{54.1} & 41.9 \\
    24 & 8 & \textbf{54.0} & 51.3 & \textbf{60.3} & 51.2 & 52.1 & \textbf{44.2} \\
    \bottomrule
  \end{tabular}
  \caption{\textbf{Balance between overview and searched frames.}
    Accuracy (\%) with Qwen3-VL-8B as the answerer and $K=32$ frames, $K_a$ from the overview and $K_d$ from the search.
    Bottom: questions grouped by the fraction of the video covered by the annotated evidence.
    V-MME: Video-MME; LongVB: LongVideoBench; CG: CG-Bench. \textbf{Bold}: best.}
  \label{tab:split}
\end{table}

\begin{figure}[t]
  \centering
  \includegraphics[width=\linewidth]{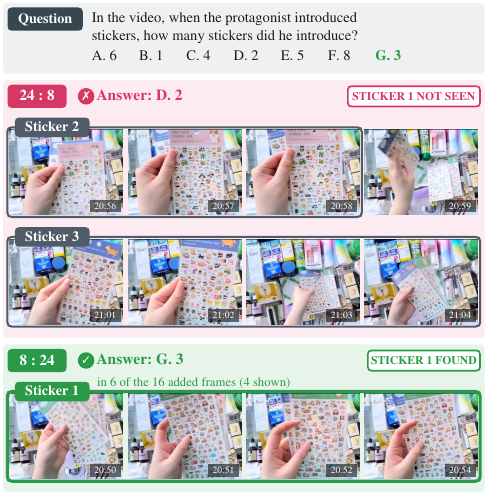}
  \caption{\textbf{An example where more searched frames help.}
    A CG-Bench question answered with $K_a{:}K_d=24{:}8$ (default) and $8{:}24$.
    No frame of $24{:}8$ shows sticker 1: its 8 searched frames show only stickers 2 and 3, and it answers 2.
    The 24 searched frames of $8{:}24$ contain the same 8 frames and 16 more, 6 of which show sticker 1, and it answers 3.
    Overview frames are not shown.}
  \label{fig:split_example}
\end{figure}

\begin{table}[t]
  \centering
  \scriptsize
  \setlength{\tabcolsep}{2pt}
  \begin{tabular}{@{}cc|@{\hspace{4pt}}cccccc@{}}
    \toprule
    $K_a$ & $K_d$ & V-MME & MLVU & LVBench & LongVB & CG & Avg. \\
    \midrule
    24 & 8 & \textbf{68.9} & \textbf{77.7} & \textbf{54.0} & \textbf{67.5} & \textbf{51.4} & \textbf{63.9} \\
    \midrule
    24 & 0 & 67.0\,{\tiny($-$1.9)} & 76.5\,{\tiny($-$1.2)} & 47.4\,{\tiny($-$6.6)} & 65.7\,{\tiny($-$1.8)} & 45.4\,{\tiny($-$6.0)} & 60.4\,{\tiny($-$3.5)} \\
    32 & 0 & 67.4\,{\tiny($-$1.5)} & 76.4\,{\tiny($-$1.3)} & 47.2\,{\tiny($-$6.8)} & 66.6\,{\tiny($-$0.9)} & 45.9\,{\tiny($-$5.5)} & 60.7\,{\tiny($-$3.2)} \\
    \midrule
    24 (Uni) & 8 & 67.6\,{\tiny($-$1.3)} & 72.1\,{\tiny($-$5.6)} & 49.8\,{\tiny($-$4.2)} & \textbf{67.5}\,{\tiny(0.0)} & 50.1\,{\tiny($-$1.3)} & 61.4\,{\tiny($-$2.5)} \\
    \bottomrule
  \end{tabular}
  \caption{\textbf{Detailed search and preliminary selection.}
    Accuracy (\%) with Qwen3-VL-8B as the answerer. First row: the default.
    (Uni): the $K_a$ frames are sampled uniformly instead of picked by the \Verifier. \textbf{Bold}: best, ties included.}
  \label{tab:components}
\end{table}

\begin{table}[t]
  \centering
  \scriptsize
  \setlength{\tabcolsep}{2pt}
  \colorlet{singlepass}{black!55}
  \begin{tabular}{@{}lccccccc@{}}
    \toprule
    Method & Frames & V-MME & MLVU & LVBench & LongVB & CG & Avg. \\
    \midrule
    WFS-SB~\cite{chen2026wfssb} & 16 & 65.0 & 70.5 & 50.7 & 64.3 & -- & -- \\
    VideoPanels~\cite{doorenbos2026videopanels} & 16 & 63.8 & 62.3 & 39.7 & 58.4 & -- & -- \\
    VideoBrain$^{\diamond}$~\cite{zou2026videobrain} & 22 & -- & -- & 41.3 & 53.3 & -- & -- \\
    \midrule
    Bayesian-GRPO~\cite{chang2026labelpriors} & 128 & 67.2 & 77.7 & -- & 63.9 & -- & -- \\
    LoHi-SemDiv~\cite{dong2026lohi} & 128 & 67.1 & 75.3 & 46.8 & -- & -- & -- \\
    QCA~\cite{peng2026qca} & 128 & -- & 76.8 & 53.2 & 67.8 & -- & -- \\
    VTS$^{\diamond}$~\cite{zhang2026vts} & 274--328 & 67.5 & 66.1 & 54.7 & 58.5 & 51.5 & 59.7 \\
    \midrule
    \method{} & 16 & 67.6 & 74.2 & 51.4 & 67.0 & 49.6 & 62.0 \\
    \quad\textcolor{singlepass}{Uniform} & \textcolor{singlepass}{16} & \textcolor{singlepass}{60.1} & \textcolor{singlepass}{59.6} & \textcolor{singlepass}{37.4} & \textcolor{singlepass}{56.2} & \textcolor{singlepass}{39.7} & \textcolor{singlepass}{50.6} \\
    \method{} & 32 & 68.9 & 77.7 & 54.0 & 67.5 & 51.4 & 63.9 \\
    \quad\textcolor{singlepass}{Uniform} & \textcolor{singlepass}{32} & \textcolor{singlepass}{63.9} & \textcolor{singlepass}{63.7} & \textcolor{singlepass}{39.1} & \textcolor{singlepass}{60.4} & \textcolor{singlepass}{40.7} & \textcolor{singlepass}{53.5} \\
    \method{} & 64 & \underline{69.0} & \underline{79.8} & \underline{55.6} & \underline{69.4} & \underline{52.9} & \underline{65.4} \\
    \quad\textcolor{singlepass}{Uniform} & \textcolor{singlepass}{64} & \textcolor{singlepass}{66.5} & \textcolor{singlepass}{68.4} & \textcolor{singlepass}{43.8} & \textcolor{singlepass}{63.1} & \textcolor{singlepass}{43.4} & \textcolor{singlepass}{57.0} \\
    \method{} & 128 & \textbf{70.4} & \textbf{80.0} & \textbf{58.2} & \textbf{70.5} & \textbf{55.3} & \textbf{66.9} \\
    \quad\textcolor{singlepass}{Uniform} & \textcolor{singlepass}{128} & \textcolor{singlepass}{68.8} & \textcolor{singlepass}{72.9} & \textcolor{singlepass}{45.9} & \textcolor{singlepass}{65.1} & \textcolor{singlepass}{46.4} & \textcolor{singlepass}{59.8} \\
    \bottomrule
  \end{tabular}
  \caption{\textbf{Frame budget.} Accuracy (\%) with Qwen3-VL-8B as the answerer.
    Gray, indented: uniform sampling.
    Frames: number of frames that the answerer reads (for VTS and VideoBrain, the average per question, which varies with the benchmark).
    WFS-SB and VideoPanels: re-run by ReUnit~\cite{tang2026reunit}, VideoPanels with 16 images that each tile several frames.
    $^{\diamond}$: answers with a model fine-tuned from Qwen3-VL-8B. --: not reported. V-MME: Video-MME; LongVB: LongVideoBench; CG: CG-Bench.
    \textbf{Bold}: best, ties included; \underline{Underline}: second best.}
  \label{tab:frames}
\end{table}

\noindent
\textbf{Balance between overview and searched frames.}
With $K=32$, \cref{tab:split} shifts frames between the answer overview and the search, with $K_a{:}K_d$ at $8{:}24$, $16{:}16$, and $24{:}8$ (the default).
The overview frames span the whole video, while the searched frames come from the windows the search narrows down to and lie in the evidence more often (\cref{tab:evidence}).
Questions about the whole video and about a short moment may therefore favor different splits.
The bottom of \cref{tab:split} groups the questions of LVBench and CG-Bench by the share of the video the evidence covers: moving to $24{:}8$ raises accuracy when the evidence covers at least 10\% of the video, lowers it at 1--10\%, and changes it little below 1\%.
Questions that need much of the video thus gain more from overview frames, and questions about details from searched frames (\cref{fig:split_example}).
Below 1\%, however, the search itself rarely reaches the evidence.
The default $24{:}8$ has the highest average, 63.9.

\smallskip
\noindent
\textbf{Gain from detailed searching.}
In the $24{:}0$ and $32{:}0$ rows of \cref{tab:components}, System~II reads only the answer overview $F_{a}$, with $K_a=24$ as in the default or $K_a=32$ to keep $K=32$.
LVBench and CG-Bench, whose annotated evidence is short (median under 1\% of the video), drop the most, from 54.0 to 47.4 and 47.2 and from 51.4 to 45.4 and 45.9.
The 8 extra overview frames of $32{:}0$ raise the average only from 60.4 to 60.7, whereas the 8 searched frames of $24{:}8$ raise it to 63.9. The searched frames lie in the evidence two to six times as often as the overview frames (\cref{tab:evidence}).
The gain of the search thus comes from where it places the frames, not from the number of frames.

\smallskip
\noindent
\textbf{Gain from preliminary selection.}
In the (Uni) row of \cref{tab:components}, the $K_a=24$ answer-overview frames are sampled uniformly instead of picked by the \Verifier from the $K_o=96$ overview frames.
The average falls from 63.9 to 61.4, most on MLVU and LVBench, by 5.6 and 4.2 points.
The picked frames lie in the evidence 3.0 and 1.2 times as often as uniform frames on CG-Bench and LVBench (\cref{tab:evidence}).
With uniform overview frames, the search still raises the average from 53.5 with 32 uniform frames to 61.4.

\smallskip
\noindent
\textbf{Frame budget.}
\cref{tab:frames} varies the number of frames $K$ that the answerer reads from 16 to 128, with Qwen3-VL-8B as the answerer.
For \method{}, we keep System~I and the action budget $B$ and scale $K_o$, $K_a$, $K_d$, and $m$ with $K$, to 384, 96, 32, and 16 for $K=128$.
From 16 to 32, 64, and 128 frames, \method{} improves on all five benchmarks, and its average rises from 62.0 to 63.9, 65.4, and 66.9.
Uniform sampling gains more from the added frames (from 50.6 to 53.5, 57.0, and 59.8), but \method{} with 32 frames remains above uniform sampling with 128 frames on all five benchmarks, by 4.1 points on average.
Even with 16 frames, its average is 2.2 points above that of uniform sampling with 128 frames, and it is higher on every benchmark except Video-MME.
With 16 frames, \method{} also beats WFS-SB, VideoPanels, and VideoBrain on every benchmark that they report, by 2.4, 9.0, and 11.9 points on average.
With 32 frames, its average is also 4.2 points above that of VTS, which reads 274--328 frames per question.

\section{Conclusion}
\label{sec:conclusion}

We presented \method{}, a System One--System Two framework for long video understanding that decouples evidence search from answer generation.
A non-thinking System~I, trained on the evidence System~II relies on, searches a tree of video windows with four role agents that make typed decisions without generating text, and a frozen System~II answers once from the frames System~I selects.
Extensive experiments on five long-video benchmarks show that \method{} consistently improves over uniform sampling with five System~II models and achieves the best average accuracy among frame selection and agentic methods, while taking much less GPU time per question than the latter.
We hope this work inspires further research on this direction.

{
    \small
    \bibliographystyle{ieeenat_fullname}
    \bibliography{main}
}

\clearpage
\maketitlesupplementary
\appendix

This supplementary material gives implementation details (\cref{sec:supp_setup}), the data (\cref{sec:supp_data}), the baselines (\cref{sec:supp_baselines}), search statistics (\cref{sec:supp_search_stats}), and qualitative examples (\cref{sec:supp_qual}).

\section{Implementation details}
\label{sec:supp_setup}

\subsection{Training}
\label{sec:supp_training}

\noindent
\textbf{Training details.}
Only the three LoRA adapters are trained, while OneJev-4B and System~II stay frozen.
Each stage trains its adapters for up to one epoch with the settings in \cref{tab:supp_hparams}, and stages (ii) and (iv) continue from the adapters of the stage before.
We keep the snapshot with the best decision accuracy, class-balanced for the \Verifier, on held-out questions: 700 from CG-Bench outside the mini set, 607 from LV-Haystack, and 300 from VideoITG.
The annotated evidence intervals are padded by 3\,s on each side.
Training, including the rollouts with System~II, is conducted on 8 NVIDIA H200 GPUs.

\begin{table}[t]
  \centering
  \footnotesize
  \begin{tabular}{@{}ll@{}}
    \toprule
    Config & Value \\
    \midrule
    Adapter & LoRA, rank 16, $\alpha=32$ \\
    LoRA modules & attention, linear attention, MLP \\
    Optimizer & AdamW, weight decay 0 \\
    Learning rate & $10^{-4}$, cosine decay to $10^{-5}$ \\
    Warmup steps & 25 \\
    Batch size & 32 decisions \\
    Gradient clipping & 1.0 \\
    Precision & bfloat16 \\
    \midrule
    Overview frames $K_o=K_a$ & 16, uniform \\
    Kept frames $K_d$ & 16 \\
    \Locator context frames per call & 16 \\
    Action budget $B$ & 24 \\
    \midrule
    DAgger rounds & 1, \Controller only \\
    DAgger expert probability $\beta$ & 0.3 \\
    Forced \Locator calls & probability 0.3, at most 2 per question \\
    \bottomrule
  \end{tabular}
  \caption{\textbf{Training hyperparameters of System~I.}}
  \label{tab:supp_hparams}
\end{table}

\smallskip
\noindent
\textbf{Windows without evidence.}
In stages (ii) to (iv), each \Locator call, at the root, for \aSwitch, or for \aRefine, is redirected with probability 0.3, at most twice per question, to the most probable candidate whose inspection range does not meet $E$, if the \Locator gives it a probability of at least 0.05.
The \Controller then inspects this window once.
The forced decisions are not used for training, but the states after them are: the \Verifier learns that the evidence lies before or after the window and that the inspected frames do not show it, and the \Controller learns the expert's action, which leaves the window.

\smallskip
\noindent
\textbf{DAgger.}
We run one round of DAgger for the \Controller and keep $\theta_g$ and $\theta_v$ from stage (ii).
The \Controller of stage (iii) searches, and at each step the expert's action is executed instead with probability $\beta=0.3$.
The expert labels the states that the search reaches, and we fine-tune $\theta_c$ on them together with the states of stage (iii).

\subsection{GPU time measurement}
\label{sec:supp_timing}

\noindent
\textbf{Setup.}
All methods in \cref{tab:efficiency} run on one NVIDIA H200 NVL GPU with batch size 1, on the same 1,000 MLVU dev questions and the same 1,000 LVBench questions, sampled at random with a fixed seed.
VTS, the slowest method, is timed on the first 218 of these LVBench questions.
We count only GPU time, so CPU video decoding and preprocessing are excluded.
Search covers all work before the answer, including the vision encoding of the frames that the search reads, and Total adds the answer.

\smallskip
\noindent
\textbf{\method{}.}
Each LoRA adapter is merged into its own copy of OneJev-4B, so every decision of System~I is one forward pass of a 4B model.
The search time includes the vision encoder that encodes the frames System~I reads, and the answer is one call of the frozen Qwen3-VL-8B on the 32 selected frames.

\smallskip
\noindent
\textbf{Baselines.}
For VideoMind-7B, the search consists of its grounder and verifier, with the planner turned off.
For VTS-8B, it consists of the CLIP-based scene split, one caption for each new segment, and every decision before the final answer.
Conan-7B answers in a single call on 32 uniform frames and generates about 700 tokens of per-frame scores and reasoning. It requests no additional frames in our runs, so we report only its total time.
ReFoCUS-1.3B, a frame selector, selects 32 of 512 frames with the 64 candidate frame sets of its released code, and we report only its search time.

\subsection{Inference}
\label{sec:supp_inference}

\noindent
\textbf{Search settings.}
At test time, every window has $N=4$ children.
System~I samples $K_o=96$ overview frames, gives the $K_a=24$ with the highest evidence probability to System~II, and keeps at most $K_d=8$ frames during a search of at most $B=24$ actions.
Each \aInspect draws $M=32$ candidate frames from the active window extended by $\mu=1/8$ of its length on each side, and the \Selector picks $m=4$ of them.
The \Verifier keeps a frame when its evidence probability is at least $\kappa=0.5$, and it judges the evidence position with $\rho=1/4$ from 32 frames of $\tilde w_t$.
Deeper in the tree, the \Locator gets one context frame per 10\,s of the parent window, at least 16 and at most 64 per call.
When the search ends with fewer than $K_d$ kept frames, the free slots are filled first from the rejected frames and then from a grid of 128 uniform positions over the video, each time with the position farthest from the frames already chosen.
Besides the frame counts $K_o$, $K_a$, and $K_d$ (\cref{tab:supp_hparams}), three settings differ from training: there, $m=8$, the \Locator gets 16 context frames per call, and the \Verifier judges the evidence position from 16 frames.

\smallskip
\noindent
\textbf{Input of System~II.}
System~II reads the $K=32$ frames in temporal order as separate images, each resized to a longer side of 640 pixels and preceded by its timestamp (\cref{fig:supp_prompt}).
It decodes greedily, with at most 64 new tokens.
If its output contains no option letter, we take the option whose letter has the highest probability at the first generated position.

\begin{figure}[t]
  \centering
  \fcolorbox{black!45}{black!4}{%
  \begin{minipage}{0.94\linewidth}
    \footnotesize\ttfamily\raggedright
    You are given frames from a \textit{\textless duration\textgreater}s video.\\
    Frame at \textit{\textless t$_1$\textgreater}s: \textit{\textless image\textgreater}\\
    Frame at \textit{\textless t$_2$\textgreater}s: \textit{\textless image\textgreater}\\
    \dots\\
    Frame at \textit{\textless t$_{32}$\textgreater}s: \textit{\textless image\textgreater}\\
    Question: \textit{\textless question\textgreater}\\
    Options:\\
    (A) \textit{\textless option A\textgreater}\\
    (B) \textit{\textless option B\textgreater}\\
    \dots\\
    Answer with the option's letter from the given choices directly.
  \end{minipage}}
  \caption{\textbf{Input of System~II.} Italic fields are filled in for each question; each \textit{\textless image\textgreater} is one frame, and timestamps have one decimal.}
  \label{fig:supp_prompt}
\end{figure}

\section{Data}
\label{sec:supp_data}

\subsection{Benchmarks}
\label{sec:supp_benchmarks}

\begin{itemize}
  \item \textbf{Video-MME}~\cite{fu2025videomme} has 2,700 questions on 900 videos, 300 each of short (under 2\,min), medium (4--15\,min), and long (30--60\,min) videos. We use it without subtitles.
  \item \textbf{MLVU}~\cite{zhou2025mlvu}: we use the 2,174 multiple-choice questions of the dev set, on 1,122 videos with a median length of 8\,min, and report M-Avg, the mean accuracy over its seven tasks.
  \item \textbf{LongVideoBench}~\cite{wu2024longvideobench}: we use the 1,337 questions of the validation set, on 753 videos of up to one hour, without subtitles.
  \item \textbf{LVBench}~\cite{wang2025lvbench} has 1,549 questions on 103 videos of 30 to 140\,min (mean 67\,min). Each question has one or more of six types and a time reference to the part of the video that answers it.
  \item \textbf{CG-Bench}~\cite{chen2025cgbench} mini has 3,000 questions on 1,118 videos of 9 to 105\,min (median 25\,min), with 5 to 8 options and annotated clue intervals.
\end{itemize}
The evidence hit rate (\cref{tab:evidence}) uses the clue intervals of CG-Bench and the time references of LVBench.

\subsection{Training data}
\label{sec:supp_train_data}

\noindent
\textbf{Sources.}
The long videos come from CG-Bench outside the mini set, whose clue intervals mark the evidence, and from the Ego4D part of LV-Haystack~\cite{ye2025tstar}, whose annotated keyframes we widen by 2\,s on each side.
The short videos are clips of the VideoITG data~\cite{wang2026videoitg} from LLaVA-Video-178K~\cite{zhang2025llavavideo}, whose annotation marks the seconds that show the evidence; each marked second $k$ becomes the span $[k-0.5,k+0.5]$.
Touching spans are merged, and the evidence set $E$ pads them by 3\,s on each side.
After filtering, the training set has 3,023 long-video questions on 1,432 videos (median length 23.5\,min) and 5,025 short-video questions on 4,261 clips (median length 78\,s).

\smallskip
\noindent
\textbf{Filter.}
We keep the questions whose evidence System~II needs (\cref{sec:method_training}), with the four stages of VTS and Qwen3-VL-8B as System~II, decoding greedily.
(1)~After spans at most 6\,s apart are merged, the evidence is one segment that covers at most 30\% of the video; the VideoITG clips may have several segments.
(2)~System~II answers incorrectly from the question and options alone.
(3)~It answers correctly from frames sampled at 1\,fps inside the evidence, at most 256.
(4)~It answers incorrectly from frames sampled at 1\,fps outside the evidence, at most 256.
\cref{tab:supp_filter} gives the number of questions left after each stage.
The held-out questions for snapshot selection (\cref{sec:supp_training}) are not filtered.

\begin{table}[t]
  \centering
  \footnotesize
  \setlength{\tabcolsep}{4pt}
  \begin{tabular}{@{}lrrrrr@{}}
    \toprule
    Source & Pool & (1) & (2) & (3) & (4) \\
    \midrule
    CG-Bench & 8,429 & 8,187 & 6,096 & 3,140 & 1,530 \\
    LV-Haystack & 11,024 & 10,925 & 5,802 & 4,173 & 1,493 \\
    VideoITG & 84,139 & 77,782 & 28,907 & 21,934 & 5,025 \\
    \bottomrule
  \end{tabular}
  \caption{\textbf{Training questions left after each stage of the filter.} Pool: the questions of each source outside the held-out sets.}
  \label{tab:supp_filter}
\end{table}

\section{Baselines}
\label{sec:supp_baselines}

\cref{tab:main} groups the baselines by the answerer that reads their frames.
For each answerer, it includes methods that report results with that answerer and at most 32 frames; VTS, which reads more frames, is compared in \cref{tab:frames}.
Single-pass methods ($^{\dagger}$) pick all frames at once, and the other methods search over several rounds.
Methods marked $^{\diamond}$ answer with a model that they fine-tune from the answerer, and they are listed in the group of that answerer.

\smallskip
\noindent
\textbf{Reported and reproduced results.}
For each method, we use only the answerers that its paper reports.
With each of them, we take the numbers that the paper reports under our protocol with at most 32 frames, and we run the released code and weights on the benchmarks that the paper does not report or reports under another protocol.
We do not evaluate a method with an answerer that its paper does not report.
Of the 110 baseline cells in \cref{tab:main}, 57 are taken from the papers, 44 come from our runs, and 9 are left empty.
T* reports Qwen2.5-VL-7B only with 8 and 16 frames, so we run it on all five benchmarks with 32 frames.

\smallskip
\noindent
\textbf{Settings of our runs.}
We use the official code, weights, and default settings.
All MLVU results are M-Avg on the dev split, and we re-evaluate the methods whose papers report the test split or a per-question average.
With Qwen2.5-VL-7B and InternVL3-8B, we keep the roles of the released VideoMind-7B, LoRA adapters on Qwen2-VL, and replace only the answerer.
The LVBench cells of these two rows come from its paper, where all roles are trained on the respective base model.
For Video-MTR, we use its 32-frame weights and take the paper numbers from its 32-frame results, while its main results use 80 frames.
For VTS, we take Video-MME and LVBench from its paper and run its released code and fine-tuned answerer on MLVU dev, LongVideoBench, and CG-Bench mini, since its paper reports MLVU on the test split and CG-Bench on a subset of 1,176 mini questions.
The uniform rows use lmms-eval with its default settings and 32 frames, and InternVL3-8B reads each frame as at most 2 tiles and a thumbnail.

\smallskip
\noindent
\textbf{Aggregation and missing results.}
When a paper reports Video-MME or LongVideoBench only by video duration, we average the duration groups, weighted by their numbers of questions.
A cell is left empty when the paper does not report the benchmark under our protocol, \eg, MLVU only on its test split, and the code or weights are not released.

\section{Search statistics}
\label{sec:supp_search_stats}

\cref{tab:supp_search} summarizes the searches of the run in \cref{tab:evidence}.
A question takes 6.43 \Controller actions on average.
Searches go deeper on longer videos: the median depth of the deepest inspected window is 2 on Video-MME, MLVU, and LongVideoBench, 3 on CG-Bench, and 4 on LVBench, where that window spans 0.39\% of the video.
Of all searches, 96.8\% end when $K_d$ frames are kept, 2.8\% when the action budget runs out, and 0.4\% when no action is legal.
For questions with a \aSwitch or \aUp, the proportion of windows containing evidence increases from 22.8\% before the first such action to 35.0\% afterward on CG-Bench, and from 20.1\% to 31.0\% on LVBench.

\begin{table*}[t]
  \centering
  \footnotesize
  \begin{tabular}{@{}lcccccccc@{}}
    \toprule
    & Length & \multicolumn{4}{c}{Actions} & & With \aSwitch & \\
    \cmidrule(lr){3-6}
    Benchmark & (min) & \aRefine & \aSwitch & \aUp & \aInspect & All actions & or \aUp (\%) & Depth \\
    \midrule
    Video-MME & 8.1 & 2.30 & 0.45 & 0.23 & 2.96 & 5.94 & 19.3 & 2 \\
    MLVU & 8.0 & 2.24 & 0.41 & 0.11 & 2.88 & 5.64 & 18.5 & 2 \\
    LVBench & 61.1 & 4.03 & 0.74 & 0.60 & 2.79 & 8.14 & 31.2 & 4 \\
    LongVideoBench & 8.4 & 2.38 & 0.60 & 0.22 & 2.69 & 5.89 & 25.2 & 2 \\
    CG-Bench & 25.4 & 3.19 & 0.59 & 0.27 & 2.73 & 6.79 & 26.2 & 3 \\
    \midrule
    All & 17.6 & 2.79 & 0.54 & 0.27 & 2.82 & 6.43 & 23.5 & 3 \\
    \bottomrule
  \end{tabular}
  \caption{\textbf{Search statistics of \method{}}, per question, in the run of \cref{tab:evidence}. Length: median video length. Actions: \Controller actions, including the first \aRefine. Depth: median depth of the deepest inspected window, with $\ell(w_0)=0$ for the whole video.}
  \label{tab:supp_search}
\end{table*}

\section{Qualitative examples}
\label{sec:supp_qual}

\begin{figure*}[t]
  \centering
  \includegraphics[width=\linewidth]{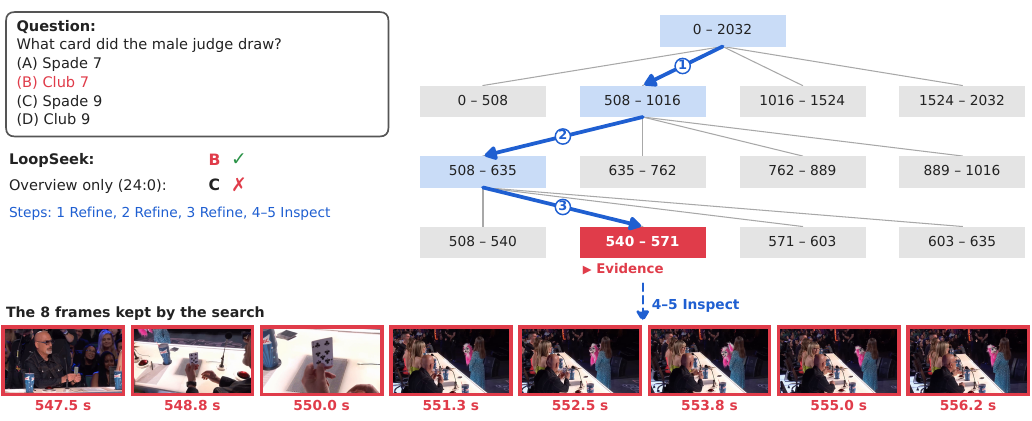}
  \caption{\textbf{Direct narrowing} (LVBench, uid 4600). The video is 2,032\,s long, and the annotated evidence lies between 547 and 560\,s. Boxes give windows in seconds: blue, visited; gray, not visited; red, the window that holds the evidence at the deepest level shown. Blue arrows: the search path, numbered by step. Bottom: the 8 frames kept by the search, with a red border when they lie in the evidence. The correct option is in red. Overview only: System~II reads only the 24 overview frames.}
  \label{fig:supp_ex_direct}
\end{figure*}

\begin{figure*}[t]
  \centering
  \includegraphics[width=\linewidth]{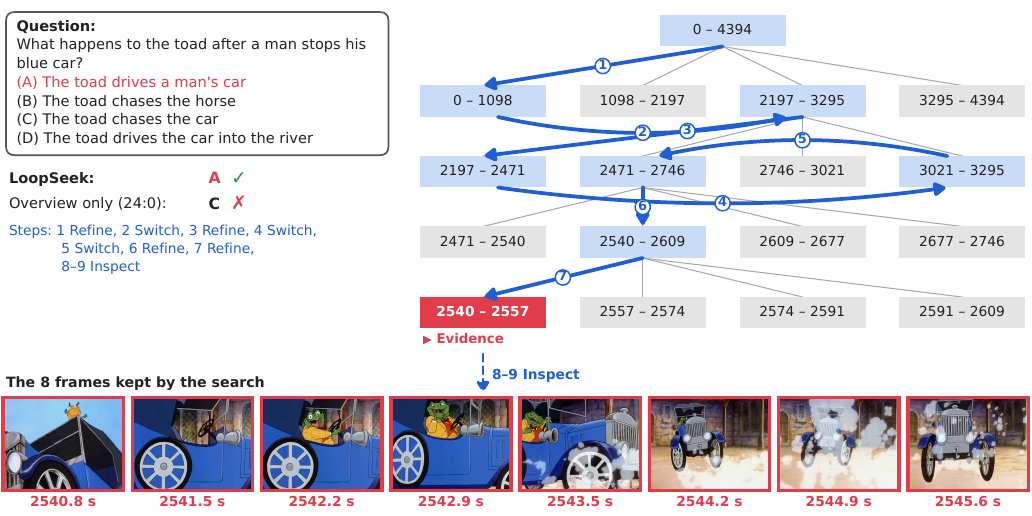}
  \caption{\textbf{Recovery with \aSwitch} (LVBench, uid 362). The video is 4,394\,s long, and the annotated evidence lies between 2531 and 2557\,s. Notation as in \cref{fig:supp_ex_direct}.}
  \label{fig:supp_ex_switch}
\end{figure*}

\begin{figure*}[t]
  \centering
  \includegraphics[width=\linewidth]{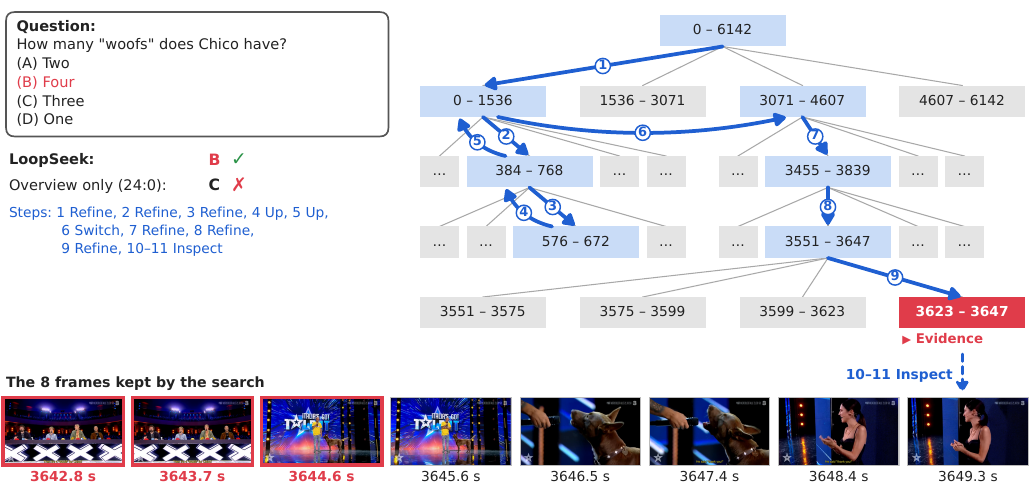}
  \caption{\textbf{Recovery with \aUp} (LVBench, uid 4705). The video is 6,142\,s long, and the annotated evidence lies between 3640 and 3645\,s. On levels with two branches, unvisited windows are gray boxes with dots. Notation as in \cref{fig:supp_ex_direct}.}
  \label{fig:supp_ex_up}
\end{figure*}

\cref{fig:supp_ex_direct,fig:supp_ex_switch,fig:supp_ex_up} show three searches on LVBench, from the run of \cref{tab:evidence}, whose time references give the annotated evidence.
Each figure shows the question, the window tree of the search, the 8 frames that the search keeps, and the answers of System~II with all 32 frames and with the 24 overview frames alone (the $24{:}0$ row of \cref{tab:components}).

\smallskip
\noindent
\textbf{Direct narrowing.}
In \cref{fig:supp_ex_direct}, three \aRefine actions narrow the 2,032\,s video to a 32\,s window that holds the evidence (547--560\,s), and two inspections keep 8 frames, all inside it.
Two of them show the card that the judge draws, the seven of clubs.
With the overview frames alone, System~II answers Spade~9.

\smallskip
\noindent
\textbf{Recovery with \aSwitch.}
In \cref{fig:supp_ex_switch}, the first window, 0--1098\,s, holds no evidence, and the \Controller switches to 2197--3295\,s.
The child that it then refines into holds no evidence either, and two more \aSwitch actions lead to 2471--2746\,s, which does.
Two \aRefine actions then reach a 17\,s window, and all 8 kept frames lie in the evidence (2531--2557\,s); they show the toad driving the car.
With the overview frames alone, System~II answers that the toad chases the car.

\smallskip
\noindent
\textbf{Recovery with \aUp.}
In \cref{fig:supp_ex_up}, three \aRefine actions go down to 576--672\,s, far from the evidence (3640--3645\,s, 0.08\% of the video).
The \Controller then takes two \aUp actions back to 0--1536\,s and switches to 3071--4607\,s, which holds the evidence; \aSwitch alone moves only among the siblings of the active window and could not leave 384--768\,s.
Three \aRefine actions then reach a 24\,s window, and two inspections keep 8 frames, 3 of them inside the evidence; two of these show the subtitle ``Then it's 4 `woofs' for Chico!''.
With the overview frames alone, System~II answers Three.

\end{document}